\documentclass[10pt,twocolumn,letterpaper]{article}

\usepackage{cvpr}
\usepackage{times}
\usepackage{epsfig}
\usepackage{graphicx}
\usepackage{amsmath}
\usepackage{amssymb}
\usepackage{multirow}
\usepackage{subfigure}

\usepackage[pagebackref=true,breaklinks=true,letterpaper=true,colorlinks,bookmarks=false]{hyperref}

\cvprfinalcopy 

\def\cvprPaperID{5518} 

\renewcommand{\vec}[1]{\mathbf{#1}}

\ifcvprfinal\fi
\begin{document}

\title{Noise-Tolerant Paradigm for Training Face Recognition CNNs}

\author{
Wei Hu$^1$ \quad Yangyu Huang$^{2*}$ \quad Fan Zhang$^1$ \quad Ruirui Li$^1$\\
$^1$Beijing University of Chemical Technology, China \quad $^2$Yunshitu Corporation, China\\     
{\tt\small $^1$\{huwei,zhangf,liruirui\}@mail.buct.edu.cn}\\
{\tt\small $^2$yangyu.huang.1990@gmail.com}\\
}

\maketitle

\begin{abstract}
   Benefit from large-scale training datasets, deep Convolutional Neural Networks(CNNs) have achieved impressive results in face recognition(FR). However, tremendous scale of datasets inevitably lead to noisy data, which obviously reduce the performance of the trained CNN models. Kicking out wrong labels from large-scale FR datasets is still very expensive, although some cleaning approaches are proposed. According to the analysis of the whole process of training CNN models supervised by angular margin based loss(AM-Loss) functions, we find that the $\theta$ distribution of training samples implicitly reflects their probability of being clean. Thus, we propose a novel training paradigm that employs the idea of weighting samples based on the above probability. Without any prior knowledge of noise, we can train high performance CNN models with large-scale FR datasets. Experiments demonstrate the effectiveness of our training paradigm. The codes are available at https://github.com/huangyangyu/NoiseFace.
\end{abstract}

\section{Introduction}

Large-scale datasets are crucial for training deep CNNs in FR, and the scale of training datasets is growing tremendously. For example, a widely used FR training dataset, MS-Celeb-1M~\cite{msceleb}, contains about 100K celebrities and 10M images. However, a previous work~\cite{imdbface} points out that a million scale FR dataset typically has a noise rate higher than 30\% (about 50\% in the original MS-Celeb-1M). The presence of noisy training data may adversely affect the final performance of trained CNNs. Though a recent work~\cite{rolnick2017deep} reports that deep CNNs still perform well even on noisy datasets containing sufficient clean data, this conclusion cannot be transferred to FR, and experiments demonstrate that noisy data apparently decrease the performance of the trained FR CNNs~\cite{imdbface}.

Large-scale datasets with high-quality label annotations are very expensive to obtain. Cleaning large-scale FR datasets with automatic or semi-automatic approaches~\cite{msceleb,lightcnn,arcface} cannot really solve this problem. As can be seen, existed large-scale FR datasets, such as MS-Celeb-1M and MegaFace~\cite{megaface}, still consist considerable incorrect labels. To obtain a noise-controlled FR dataset, manual annotation is inevitable. Although an approach is introduced to effectively build a high-quality dataset IMDB-Face~\cite{imdbface}, it actually further demonstrates the difficulties of obtaining a large-scale well-annotated FR dataset. For example, it took 50 annotators one month to clean the IMDB-Face dataset, which only contains 59K celebrities and 1.7M images.


Numerous training approaches have been investigated aiming to train classification CNNs with noisy datasets~\cite{patrini2017making, xiao2015learning,sukhbaatar2014training,goldberger2016training,veit2017learning,han2018masking,li2017learning,ghosh2017robust}, but most of them are not suitable for training FR models, because of the special characters of FR datasets (discussed in the Section~\ref{secRelatedWorks}). Recently, weighting training samples is a promising direction~\cite{mentornet,coteaching,decoupling} to deal with noisy data. However, extra datasets or complex networks are required in these approaches, limiting the use of them in FR.

The CNN models in FR are usally trained with loss functions, which aim to maximize inter-identity variation and minimize intra-identity variation under a certain metric space. Very recently, some angular margin based loss(AM-Loss for short in the paper) functions~\cite{sphereface,arcface,cosface,asoftmax} are proposed and achieve the state-of-the-art performance.

In this paper, we propose a noise-tolerant paradigm to learn face features on a large-scale noisy dataset directly, different from other related approaches~\cite{lightcnn,arcface} which aim to clean the noisy dataset firstly. When training an AM-Loss supervised CNN model, the $\theta$ histogram distribution of training samples(the $\theta$ distribution for short, described in Section~\ref{subsecEffectofNoise}) is employed to measure the possibility that a sample is correctly labeled, and this possibility is then used to determine the training weight of the sample. Throughout the training process, the proposed paradigm can alleviate the impact of noisy data by dynamically adjusting the weight of samples, according to their $\theta$ distribution at that time.

To summarize, our major works are as follows:

\begin{enumerate}
\item We observe that the $\theta$ value of a clean sample has higher probability to be smaller than that of a noisy sample for an AM-Loss function. In other word, the possibility that a sample is clean can be dynamically reflected by its position in the $\theta$ distribution.
\item Based on the above observation, we employ the idea of weighting training samples, and present a novel noise-tolerant paradigm to train FR CNN models with a noisy dataset end-to-end. Without any prior knowledge of noise in the training dataset (noise rate, small clean sets, etc.), the models can achieve comparable, or even better performance, compared with the models trained by traditional methods with the same dataset without noisy samples.
\item Although many approaches are proposed to train classification models with noisy datasets, none of them aims to train FR CNN models. To the best of our knowledge, the proposed paradigm is the first to study the method to significantly eliminate adverse effects of extremely noisy data with deep CNN models in FR. Our paradigm is also the first to estimate the noise rate of a noisy dataset accurately in FR. Furthermore, our trained models can also achieve good performance on clean datasets too.
\end{enumerate}

\section{Related Works}
\label{secRelatedWorks}

\subsection{Training with Noisy Data}
Learning with noisy datasets has been widely explored in image classification training~\cite{frenay2014classification}. In classic image classification datasets, the real-world noisy labels exhibit multi-mode characteristics. Therefore, many approaches use pre-defined knowledge to learn the mapping between noisy and clean annotations, and focus on estimating the noise transition matrix to remove or correct mis-labeled samples~\cite{menon2015learning,liu2016classification,natarajan2013learning}.
Recently, it has also been studied in the context of deep CNNs. \cite{xiao2015learning} relies on manually labeling to estimate the matrix. \cite{sukhbaatar2014training,goldberger2016training} add layer in CNN models to learn the noise transition matrix. \cite{veit2017learning,hendrycks2018using} use a small clean dataset to learn a mapping between noisy and clean annotations. \cite{patrini2017making,ghosh2017robust} use noise-tolerant loss functions to correct noisy labels. Li \etal \cite{li2017learning} construct a knowledge graph to guide the learning process. Han \etal \cite{han2018masking} propose a human-assisted approach which incorporates an structure prior to derive a structure-aware probabilistic model. Different from the common classification datasets, FR datasets always contain a very large number of classes(persons), but each class contains relatively small number of images, making it difficult to find relationship patterns from noisy data. Furthermore, noisy labels behave more like independent random outliers in FR datasets. Therefore, the transition matrix or the relationship between noisy and clean labels is very hard to be estimated from FR datasets.

Some approaches attempt to update the trained CNNs only with separated clean samples, instead of correcting the noisy labels. A Decoupling technique~\cite{decoupling} trains two CNN models to select samples that have different predictions from these two models, but it cannot process heavy noisy datasets. Very recently, weighting training samples becomes a hot topic to learn with noisy datasets.

\subsubsection{Weighting Training Samples}
Weighting training samples is a well studied technique and can be applied to adjust the contributions of samples for training CNN models~\cite{friedman2001elements,focalloss,selfpaced}. Huber loss~\cite{friedman2001elements} reduces the contribution of hard samples by down-weighting the loss of them. In contrast, Focal loss~\cite{focalloss} adds a weighting factor to emphasize hard samples for training high accurate detector. Multiple step training is adopted in ~\cite{selfpaced} to encourage learning easier samples first.

The idea of weighting training samples is employed to train models with noisy datasets too, since clean/noisy samples are usually corresponding to easy/hard samples. The key of weighting samples is an effective method to measure the possibility that a sample is easy/clean. Based on Curriculum learning~\cite{curriculum}, MentorNet~\cite{mentornet} and Coteaching~\cite{coteaching} try to select clean samples using the small-loss strategy, but the noise level should be provided in advanced in~\cite{coteaching}, and a small clean set and a pre-training model are suggested in ~\cite{mentornet}. Ren \etal~\cite{reweight} also employ the clean validation set to help learning sample weights. Although FR can be regarded as a classification problem, its small subset/validation dataset is usually uncertain to be clean, even not available at all. To conclude, weighting samples is a promising direction to train CNN models with noisy datasets, but estimating sample weights usually requires complex techniques and extra knowledge.

\subsection{Loss functions in FR}
Deep face recognition has been one of the most active field in these years. Usually, FR is trained as a multi-class classification problem in which the CNN models are usually supervised by the softmax loss~\cite{deepface,frbyclass1,frbyclass2}. Some metric learning loss functions, such as contrastive loss~\cite{casia,contrastiveloss}, triplet loss~\cite{tripletloss,facenet} and center loss~\cite{centerloss}, are also applied to boost FR performance greatly. Other loss functions~\cite{marginalloss,rangeloss} also demonstrate effective performance on FR. Recently, some normalization~\cite{lsoftmax,normface,ringloss} and angular margin~\cite{sphereface,arcface,asoftmax,cosface} based methods are proposed and achieve outperforming performance, and get more attention.


\section{Preliminaries}
\subsection{Angular Margin based Losses}
Recently, the angular margin based loss functions, which have intrinsic consistency with Softmax loss, greatly improves the performance of FR. The traditional Softmax loss is presented as
\begin{equation}\label{equSoftmaxLoss1}
  L = -\frac{1}{N}\sum_{i=1}^{N}log\frac{e^{\vec{W}_{y_i}^T\vec{x}_i+\vec{b}_{y_i}}}{\sum_{j=1}^{C}{e^{\vec{W}_{j}^T\vec{x}_i+\vec{b}_{j}}}},
\end{equation}
where $\vec{x}_i$ denotes the feature of the $i$-th sample which belongs to the $y_i$-th class. $\vec{W}_j$ denotes the $j$-th column of the weights $\vec{W}$ in the layer and $\vec{b}$ is the bias term. $N$ and $C$ is the batch size and the class number. In all AM-Losses, the bias $\vec{b}_j$ is fixed to be 0 and $\|\vec{W}_j\|$ is set to 1, then the target logit~\cite{pereyra2017regularizing} can be reformulated as
\begin{equation}\label{equLogit}
  \vec{W}_{j}^T\vec{x}_i = \|\vec{x}_i\|cos\ \theta_{i,j},
\end{equation}
where $\theta_{i,j}$ is the angle between $\vec{W}_j$ and $\vec{x}_i$. We can further fix $\|\vec{x}_i\|=s$, and
the Softmax loss can be reformulated as
\begin{equation}\label{equSoftmaxLoss2}
  L = -\frac{1}{N}\sum_{i=1}^{N}log\frac{e^{s\ cos\ \theta_{i,y_i}}}{\sum_{j=1}^{C}{e^{s\ cos\ \theta_{i,j}}}},
\end{equation}
and we refer this loss as L2-Softmax in this paper. Actually, other AM-Losses have similar loss functions, but with slightly different decision boundaries.

In all AM-Losses, $\vec{W}_j$ can be regarded as the anchor of the $j$-th class. During training, the angle $\theta_{i,j}$ will be minimized for an input feature $\vec{x}_i$ belonging to the $j$-th class, and the angle $\theta_{i,k(k\neq j)}$ will be maximized at the same time. As discussed in ~\cite{arcface}, the $\theta_{i,j}$ of an input $\vec{x}_i$ belonging to the $j$-th class reflects the difficulty of training the corresponding sample for a full trained CNN model, and the $\theta$ distribution of all training samples can implicitly demonstrate the performance of the model. We firstly investigate the effect of noisy data on the $\theta$ distributions.

\subsection{Effect of Noise on $\theta$ Distributions}
\label{subsecEffectofNoise}

To investigate the effect of noise, WebFace-Clean\footnote{https://github.com/happynear/FaceVerification} containing 10K celebrities and 455K images is chosen as the clean dataset. It is cleaned by manually removing incorrect images from the original CASIA-WebFace~\cite{casia} containing 494K images. ~\cite{imdbface} estimates that there are about 9.3\%-13\% mis-labeled images in the original CASIA-WebFace, so we can regard WebFace-Clean as a noise-free dataset. Several experiments are performed, and their input and training settings are described in the Section~\ref{secExperiment}.

We firstly build another new noise-free FR dataset, named \textbf{WebFace-Clean-Sub}, which contains 60\% images randomly chosen from all celebrities in WebFace-Clean. The remaining 40\% images are used to synthesis noisy data, named \textbf{WebFace-Noisy-Sub}. Noise in FR datasets mainly fall into two types: label flips, where an image has been given a label of another class within the dataset, and outliers, where an image does not belong to any of the classes, but mistakenly has one of their labels, or non-faces can be found in the image. To generate Web-Noisy-Sub, we synthesize label flips noise by randomly changing face labels into incorrect classes, and simulate outlier noise by randomly polluting data with images from MegaFace~\cite{megaface}, and we keep the ratio of \textbf{label flips} and \textbf{outliers} at 50\%:50\%. Therefore, we get a new noisy dataset \textbf{WebFace-All} containing WebFace-Clean-Sub and WebFace-Noisy-Sub, and its noise rate is 40\%.

A ResNet-20 model(CNN-All-L2)~\cite{sphereface} supervised with L2-Softmax($s=32$) is trained with WebFace-All. For comparison, we also train another ResNet-20 model(CNN-Clean-L2) with WebFace-All, but with a small modification: a sample will be dropped in training if it belongs to WebFace-Noisy-Sub. Therefore, CNN-Clean-L2 can be considered to be trained only with clean samples.

In each training iteration, we use $\vec{W}_j$ to compute $\theta_{i,j}$ for all samples belonging to the $j$-th class. For simplification, only $cos\theta$ is computed in our implementation. We refer the $\cos\theta$ histogram distributions(the bin size is 0.01) of all training samples as $Hist_{all}$. Moreover, we compute the distributions of clean and noisy samples separately as $Hist_{clean}$ and $Hist_{noisy}$.

Figure~\ref{figDist1} shows the distributions of CNN-Clean-L2 and CNN-All-L2. The test accuracy on LFW~\cite{lfw} is used to demonstrate the performance of the trained CNNs. From Figure~\ref{figDist1}, we have following observations:

\begin{figure*}[!tbh]
\centering
\subfigure[CNN-Clean-L2]
{{\includegraphics[width=1.0\linewidth]{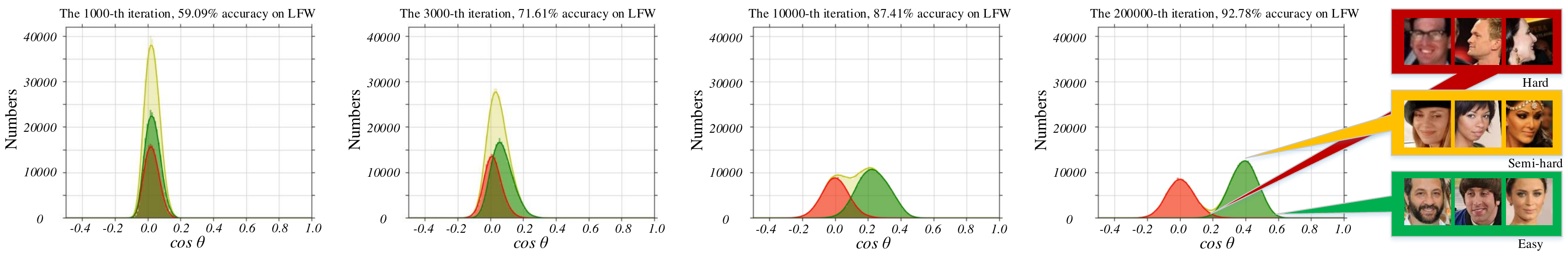}}\label{fig1a}}
\subfigure[CNN-All-L2]
{{\includegraphics[width=1.0\linewidth]{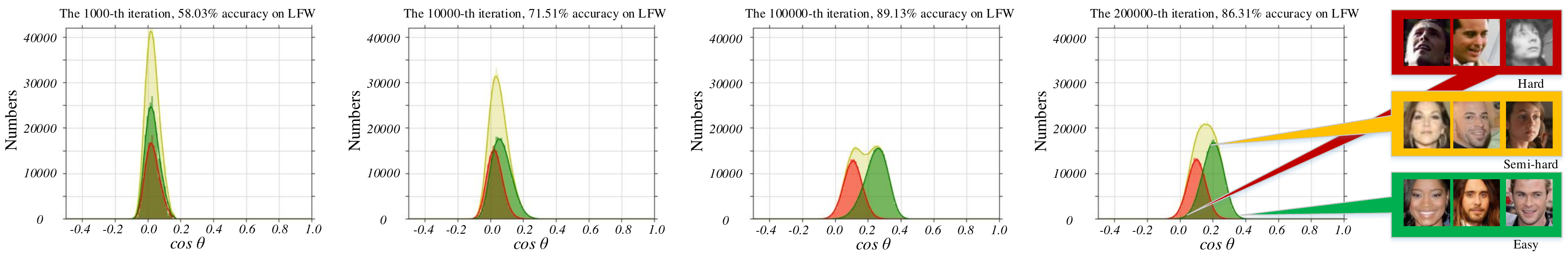}}\label{fig1b}}
\caption{The $cos\theta$ histogram distributions of CNN-Clean-L2 (top) and CNN-All-L2 (bottom). $Hist_{clean}$, $Hist_{noisy}$ and $Hist_{all}$ are colored with Green, Red and Yellow respectively. The curve edge of each distribution is smoothed with a mean filter with size = 5 to remove noise (described in the Section~\ref{secHistAll}).}
\label{figDist1}
\end{figure*}

\begin{enumerate}
\item $Hist_{clean}$ and $Hist_{noisy}$ are all Gaussian-like distributions throughout the training process for CNN-All-L2 and CNN-Clean-L2. Experiments in ArcFace~\cite{arcface} also demonstrate this phenomenon. The Gaussian-like distributions should be caused by similar quality distributions~\cite{imdbface} in FR datasets.
\item At the beginning of training,$Hist_{clean}$ and $Hist_{noisy}$ are largely overlapping, making them impossible to be separated from each other, since the CNNs are initially untrained.
\item After a short period, $Hist_{clean}$ starts to move to the right. If noisy samples are involved in training (CNN-All-L2), $Hist_{noisy}$ moves to the right too, but it has been always on the left side of $Hist_{clean}$. Therefore, the samples with larger $cos\theta$ values have larger probability to be clean. This phenomenon is mainly because that the CNNs memorize easy/clean samples quickly, and can also memorize hard/noisy samples eventually~\cite{arpit2017closer}.
\item In the latter stage of training process, the $cos\theta$ of a sample in $Hist_{clean}$ reflects the quality of the corresponding face image. Some face images of easy, semi-hard, and hard clean samples are provided in Figure~\ref{figDist1}.
\item The performance of CNN-All-L2 is adversely affected by noisy data. From the distribution in Figure~\ref{fig1b}, we can observe the negative impact in two aspects: (1) $Hist_{clean}$ and $Hist_{noisy}$ have large overlapping regions throughout the training process; (2) Compared with the $Hist_{clean}$ in Figure~\ref{fig1a}, the $Hist_{clean}$ in Figure~\ref{fig1b} is on the left side.
\end{enumerate}

\begin{figure}[!tbh]
\centering
\subfigure[60\% noisy samples]
{{\includegraphics[width=0.45\columnwidth]{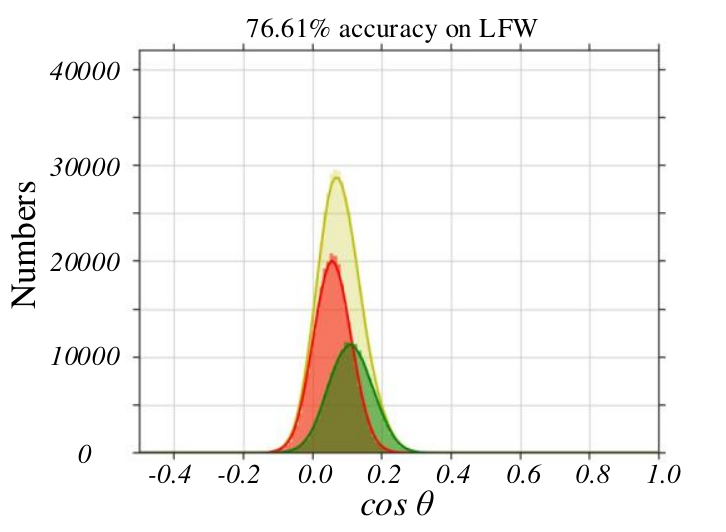}}}
\subfigure[ArcFace loss function]
{{\includegraphics[width=0.45\columnwidth]{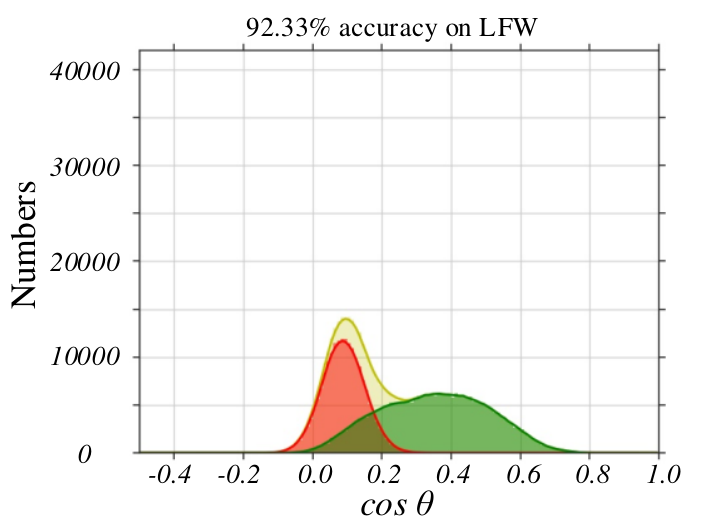}}}
\subfigure[ResNet-64 model]
{{\includegraphics[width=0.45\columnwidth]{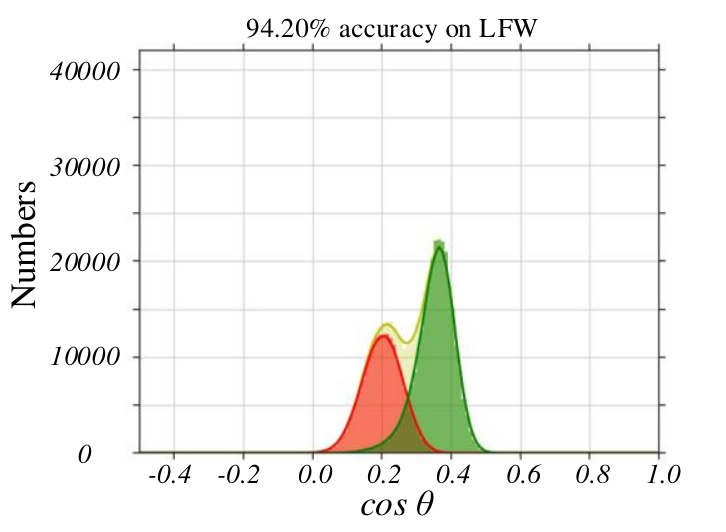}}}
\subfigure[IMDB-Face dataset]
{{\includegraphics[width=0.45\columnwidth]{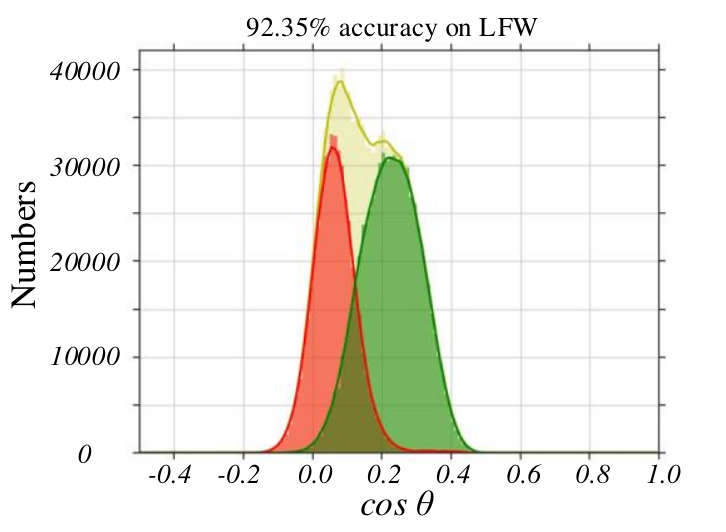}}}
\caption{The final $cos\theta$ distributions of other four models. These distributions further confirm our observations.}
\label{figDist2}
\end{figure}

These observations can be explained in theory, and more experiments are further performed to confirm them: (1)We increase the noise rate from 40\% to 60\%; (2)ArcFace~\cite{arcface} is employed to supervise the CNNs; (3) we replace the ResNet-20 with a deeper ResNet-64~\cite{sphereface}; (4) Another clean dataset IMDB-Face~\cite{imdbface}\footnote{We only downloaded 1.2M images of total 1.7M images with the provided URLs} is chosen to replace WebFace-Clean. Their final $cos\theta$ distributions shown in Figure~\ref{figDist2} further approve our observations.

CNN-Clean-L2 is actually trained by using a ideal paradigm: ignoring all noisy sample in training. However, it is difficult to predict if a sample is noisy in real training. In this paper, we propose a paradigm to minimize the impact of noisy samples based on the $cos\theta$ distributions.

\section{The Proposed Paradigm}

We propose a new training paradigm for learning face features from large-scale noisy data based on above observations. In each mini-batch training, we compute $cos\theta$ for all training samples, and the current distribution $Hist_{all}$. We define $\delta_l$ and $\delta_r$ are the leftmost/rightmost $cos\theta$ values in $Hist_{all}$. Based on the first observation in Section~\ref{subsecEffectofNoise}, no more than 2 peaks can be detected in $Hist_{all}$. $\mu_l$ and $\mu_r$ denote the $cos\theta$ values of the left/right peaks respectively, and $\mu_l=\mu_r$ if there is only one peak.

\begin{figure*}[!tbh]
\centering
\subfigure[All samples are equally treated]
{{\includegraphics[width=0.3\linewidth]{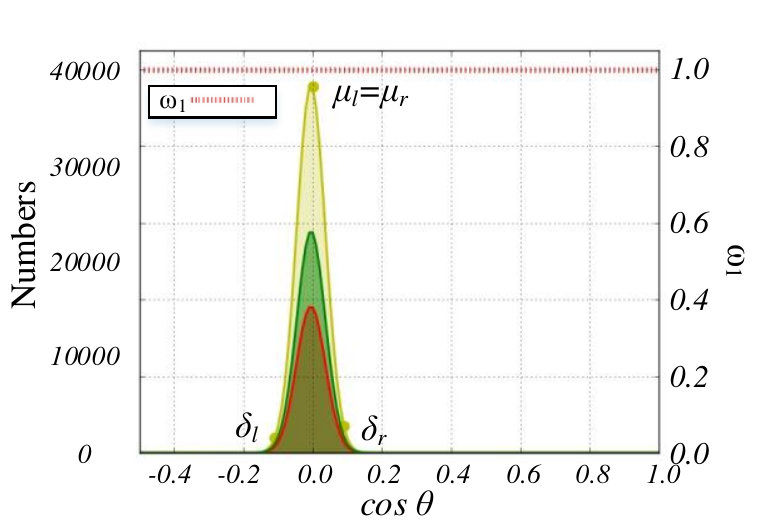}}\label{fig3a}}
\subfigure[Clean samples are emphasized]
{{\includegraphics[width=0.3\linewidth]{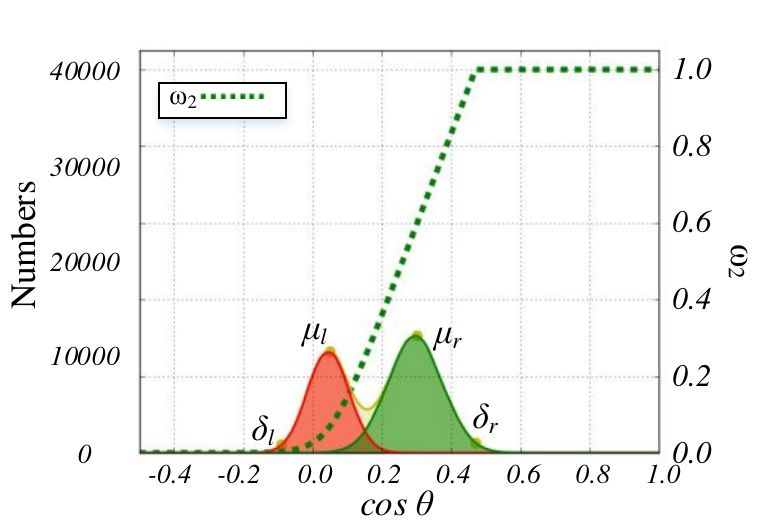}}\label{fig3b}}
\subfigure[Semi-hard clean samples are emphasized]
{{\includegraphics[width=0.3\linewidth]{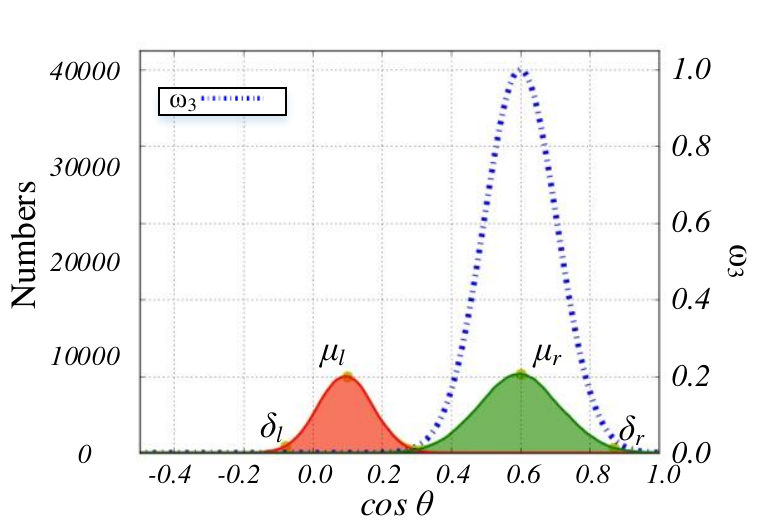}}\label{fig3c}}
\caption{The $cos\theta$ distributions and the corresponding $\omega$ of three strategies.}
\label{figDist3}
\end{figure*}

The target of the paradigm is to correctly evaluate the probability of a sample being clean in training, and then adjust the weight of the sample according to this probability. The key idea of our paradigm can be briefly introduced as:
\begin{enumerate}
\item At the beginning of training process, all samples are treated equally.

According to the 2nd observation in Section~\ref{subsecEffectofNoise}, at the beginning of training process, the trained CNN does not have the ability for face recognition. Therefore, all samples should have the same weight for training.

\item After a short period of training, samples with larger $cos\theta$ have larger weight.

According to the 3rd observation in Section~\ref{subsecEffectofNoise}, after a short period of training, samples with larger $cos\theta$ should have larger weight for training.

\item In the end of training, semi-hard clean samples are emphasized to further promote the performance.

According to the 4th observation in Section~\ref{subsecEffectofNoise}, easy, semi-hard and hard clean samples can be distinguished according to their $cos\theta$ in $Hist_{clean}$ at this time. The pre-condition of training semi-hard clean samples is that the trained CNN already has good performance, and the overlapping area of $Hist_{clean}$ and $Hist_{noisy}$ is relatively small. In this circumstance, semi-hard clean samples will have larger weight than other samples for training. Training semi-hard clean samples is also implicitly adopted in ArcFace~\cite{arcface} and FaceNet~\cite{facenet}.

\end{enumerate}

We present three corresponding strategies to compute sample weights as following:

\textbf{Strategy One} In this strategy (see Figure~\ref{fig3a}), all samples have the same weight as
\begin{equation}\label{equW1}
  \omega_{1,i} = 1,
\end{equation}
where $\omega_{1,i}$ is the weight of an input sample $\vec{x}_i$.

\textbf{Strategy Two} In this strategy (see Figure~\ref{fig3b}), the samples with larger $cos\theta$ have larger weight as
\begin{equation}\label{equW2}
  \omega_{2,i} = \frac{softplus(\lambda z)}{softplus(\lambda)},
\end{equation}
where $z=\frac{cos\theta_{i,j} - \mu_l}{\delta_r - \mu_l}$, and $softplus(x)=log(1+e^x)$ is a smooth version of the RELU activation\cite{rectified}. $\lambda$ is used to normalize the function, and $\lambda=10$ in all experiments.

\textbf{Strategy Three} In this strategy (see Figure~\ref{fig3c}), the semi-hard clean samples are emphasized. We define $\mu_r$ as the $cos\theta$ value of the right peak in $Hist_{all}$ ($\mu_l$ corresponding to the left peak), which can be consider as the center of $Hist_{clean}$ (according to the 5-th and the 2-th observation). The sample weight is computed as
\begin{equation}\label{equW3}
  \omega_{3,i} = e^{-{(cos\theta_{i,j} - \mu_r)^2} {\Huge /} {2\sigma^2}},
\end{equation}
where $\sigma^2$ is the variance of the $Hist_{clean}$, which can be approximated by using the part to the right of $\mu_r$. We set $\sigma = {(\delta_r - \mu_r)}{\huge /}{2.576}$ to cover 99\% samples in $Hist_{clean}$.

\subsection{Compute Time-Varying Fusion Weight}
\label{seqFusion}
\begin{figure*}[!tbh]
\centering
\subfigure[$\alpha(\delta_r)$, $\beta(\delta_r)$, and $\gamma(\delta_r)$]
{{\includegraphics[width=0.25\linewidth]{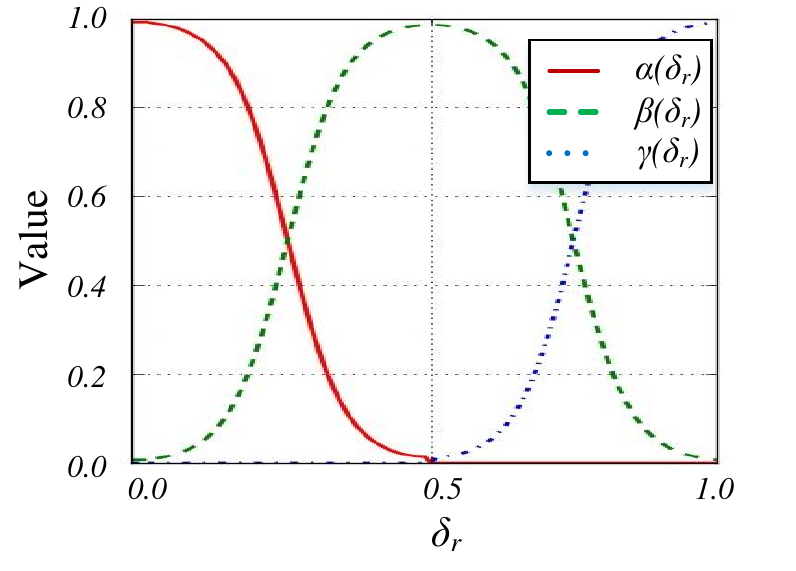}}}
\subfigure[Two fusion examples]
{{\includegraphics[width=0.6\linewidth]{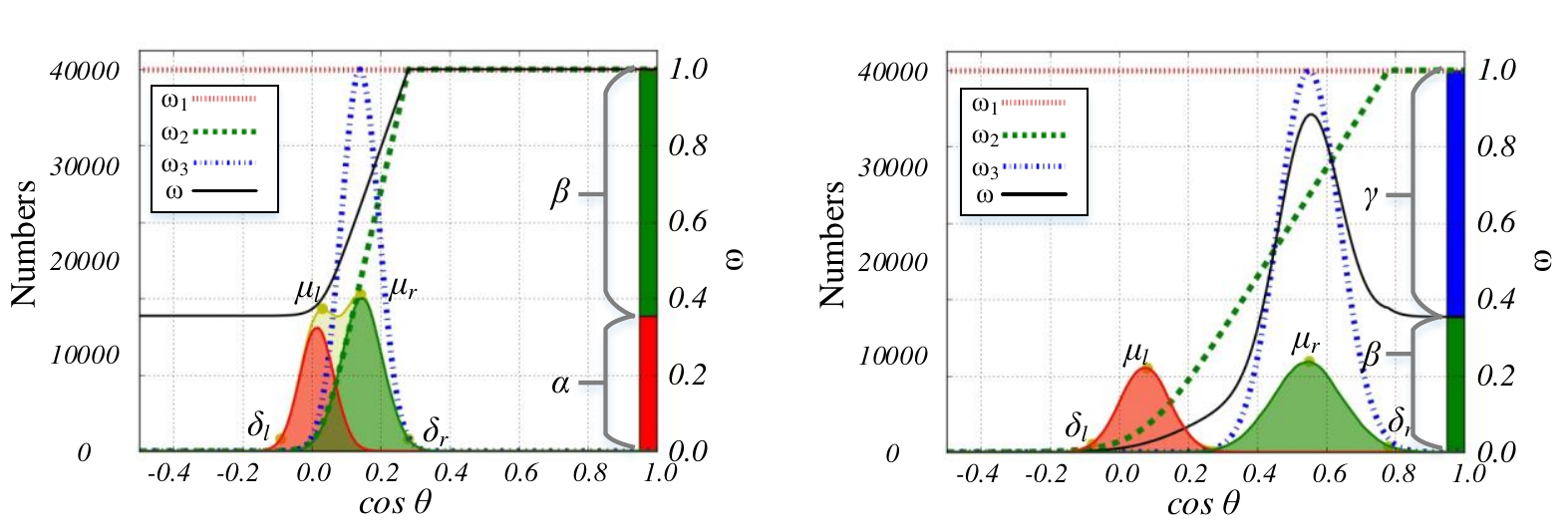}}}
\caption{The left figure demonstrates three functions: $\alpha(\delta_r)$, $\beta(\delta_r)$, and $\gamma(\delta_r)$. The right figure shows two fusion examples. According to the $\omega$, we can see that the easy/clean samples are emphasized in the first example($\delta_r<0.5$), and the semi-hard clean samples are emphasized in the second example($\delta_r>0.5$).}
\label{figFusion}
\end{figure*}

Three weighting sample strategies are introduced as above, but how to select the applied strategy in training? There is no clear criteria. Inspired by the gradually learning technique in Co-teaching~\cite{coteaching}, we compute the sample weight in a fusion way.

According to our observations, $\delta_r$ is a good signal to approximately reflect the performance of the trained CNN. As the CNN achieves better performance, $\delta_r$ will gradually move to the right. We define a threshold $\zeta$ to divide possible $\delta_r$ values into two ranges: $[0, \zeta]$ and $(\zeta,1]$. The selected strategy is changed from the 1st one to the 2nd one in the first range, and from 2nd one to the 3rd one in the second range. In all experiments, we set $\zeta=0.5$. Then, we compute the sample weight as
\begin{equation}\label{equWF}
  \omega_i = \alpha(\delta_r)\omega_{1,i}+\beta(\delta_r)\omega_{2,i}+\gamma(\delta_r)\omega_{3,i},
\end{equation}
where
\begin{equation}\label{alpha}
\alpha(\delta_r)=(2 - \frac{1}{1+e^{5-20\delta_r}} - \frac{1}{1+e^{20\delta_r-15}})\lceil0.5-\delta_r\rceil,
\end{equation}
, $\beta(\delta_r)=1-\alpha(\delta_r)-\gamma(\delta_r)$ and $\gamma(\delta_r)=\alpha(1.0 - \delta_r)$. As shown in Figure~\ref{figFusion}, at first, $\omega_{1,i}$ has the greatest impact. As $\delta_r$ gradually moves to right, $\omega_{2,i}$ and then $\omega_{3,i}$ begin to play more important roles.

\subsubsection{Discussion on Weight Computation}

Equation~\ref{equW1}, Equation~\ref{equW2} and Equation~\ref{equW3} are used to compute the sample weight during training. These 3 equations directly reflect the key idea of our paradigm, but they are only introduced empirically and have some free parameters. The same situation also exists in Equation~\ref{equWF}. The key contribution of our approach is the ideas of training paradigm and weight fusion. We also perform some experiments, and found that good performance can be achieved as long as the used equations can correctly reflect the key ideas. Therefore, these equations are provided for reference, and more exploration could be conducted to find other theoretical justified equations.

\subsection{Implementation Details}

\subsubsection{$Hist_{all}$ Related Variables}
\label{secHistAll}
According to the Equation~\ref{equWF}, $cos\theta_{i,j}$, $\delta_r$, $\delta_l$ and $\mu_r$ are required to compute the final weight of the sample $\vec{x}_i$ belonging to the $j$-th class. Except $cos\theta_{i,j}$, other variables are computed based on $Hist_{all}$.

In theory, $Hist_{all}$ should be computed in each mini-batch training, which is very time-consuming because the number of samples is usually very large in FR datasets. In our implementation, the $cos\theta$ values of recent $K$ training samples are stored to compute another distribution $Hist_{K}$. The training samples are pre-shuffled, $Hist_{K}$ can be considered as an approximate $Hist_{all}$ with a suitable $K$. We set $K=64,000$ (1000 batches) in our experiments.

To resist noise, a mean filter with size 5 is firstly applied to $Hist_{K}$ to remove noise. We select the top 0.5\% leftmost/rightmost $cos\theta$ values as $\delta_l$ and $\delta_r$. A very simple method is applied to find all peaks in $Hist_K$: the number of frequency in a bin is larger than all of its left/right neighbour bins (Radius = 5). Theoretically, we can only find one or two peaks during training. However, we sometimes find more than two peaks, or find no peak at all, because $Hist_{clean}$ and $Hist_{noisy}$ are actually not always Gaussian-like distributions. We employ a simple technique to find $\mu_r$: if there is only one peak $\in (\zeta,1]$, its $cos\theta$ is the $\mu_r$, and if more than one peaks are found, we choose the highest one. Similar method can be used to find $\mu_l$.

When the noise rate is very high, $\mu_r$ may become difficult to detect. In this circumstance, the key is to train easy/clean samples as much as possible, and $\omega_{2,i}$ should play more important role than $\omega_{3,i}$. Therefore, missing $\mu_r$ may have few impact on the final performance. In the contrary, if the noise rate is very low, missing $\mu_l$ may also have few impact on the final performance.

\begin{figure*}[tbh]
\centering
{\includegraphics[width=0.95\linewidth]{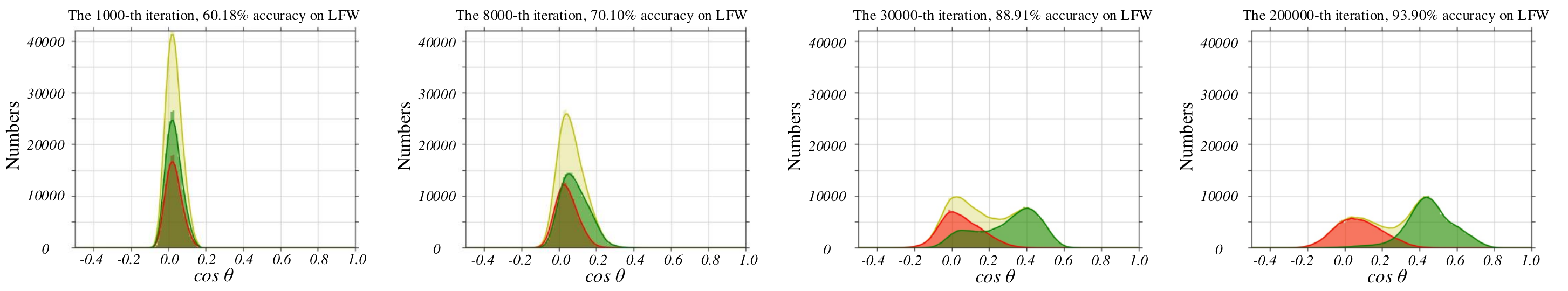}}
\caption{The $cos\theta$ distributions of our CNN model trained with 40\% noisy samples.}
\label{figDist4}
\end{figure*}

\subsubsection{Weighting in AM-Losses}
\label{secWL}
\textbf{Method 1} Usually, the weight is applied to minimize the loss
and the weighted loss function of L2-Softmax is
\begin{equation}\label{equSoftmaxLoss2_F1}
L_1 = -\frac{1}{N}\sum_{i=1}^{N}\omega_i log\frac{e^{s\ cos\ \theta_{i,y_i}}}{\sum_{j=1}^{C}{e^{s\ cos\ \theta_{i,j}}}}.
\end{equation}
Moreover, there is another method to apply sample weights.

\textbf{Method 2} In AM-Losses, an input $\vec{x}_i$ is normalized and re-scaled with a parameter $s$ ($\|\vec{x}_i\|$ can be regarded as $s$ in SphereFace). The scaling parameter $s$ is better to be a properly large value as the hypersphere radius, according to the discussion in ~\cite{cosface,arcface}. A small $s$ can lead to insufficient convergence even no convergence. The lower bound of $s$ is also discussed in ~\cite{cosface,l2scale}. Inspired by the effect of $s$, we can also apply the weights $\omega_i$ to the scaling parameter during training CNNs. Therefore,
the loss function of L2-Softmax can be formulated as
\begin{equation}\label{equSoftmaxLoss2_F2}
  L_2 = -\frac{1}{N}\sum_{i=1}^{N}log\frac{e^{\omega_i s\ cos\ \theta_{i,y_i}}}{\sum_{j=1}^{C}{e^{\omega_i s\ cos\ \theta_{i,j}}}}.
\end{equation}

Similar method can be applied in other AM-Losses too. Two methods all can be employed to train CNNs with noisy datasets. According to our experiments, the latter one shows better performance in most cases.

\section{Experiments}
\label{secExperiment}
To verify the effectiveness of our method, several experiments are performed. In these experiments, face images and landmarks are detected by MTCNN~\cite{mtcnn}, then aligned by similar transformation as~\cite{lightcnn}, and cropped to $128 \times 128$ RGB images. Each pixel in RGB images is normalized by subtracting 127.5 then dividing by 128. We use Caffe~\cite{caffe} to implement CNN models. For fair comparison, all CNN models are trained with SGD algorithm with the batch size of 64 on 1 TitanX GPUs. The weight decay is set to 0.00005. The learning rate is initially 0.1 and divided by 10 at the 80K, 160K iterations, and we finish the training process at 200K iterations.

First, we perform a similar experiment as in the Section~\ref{subsecEffectofNoise}, but using the proposed training paradigm. Figure~\ref{figDist4} shows the $cos\theta$ distributions during training. It is obvious that $Hist_{clean}$ are separated from $Hist_{noisy}$. According to the final distributions in Figure~\ref{figDist4}, Figure~\ref{fig1a} and Figure~\ref{fig1b}, the adverse effect from noisy samples is largely eliminated this time.

\begin{figure}[tbh]
\centering
\subfigure[60\% noisy samples]
{{\includegraphics[width=0.45\columnwidth]{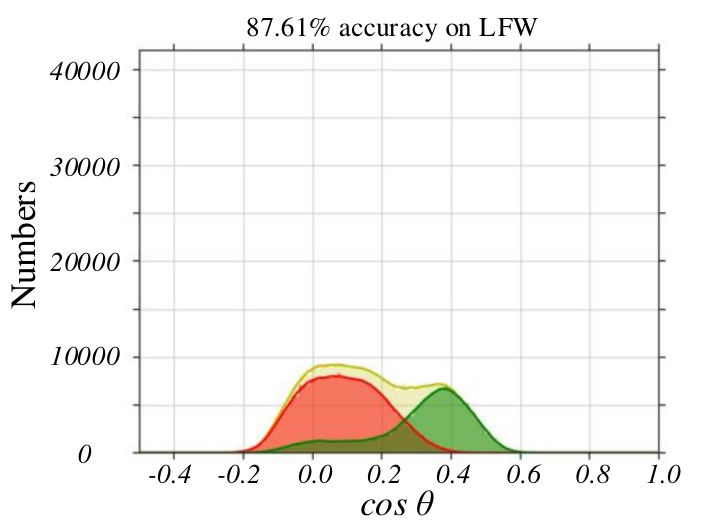}}}
\subfigure[ArcFace loss function]
{{\includegraphics[width=0.45\columnwidth]{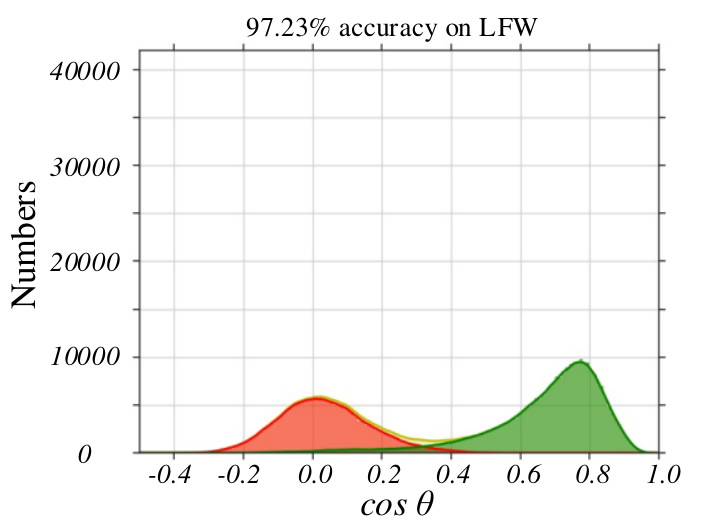}}}
\subfigure[ResNet-64 model]
{{\includegraphics[width=0.45\columnwidth]{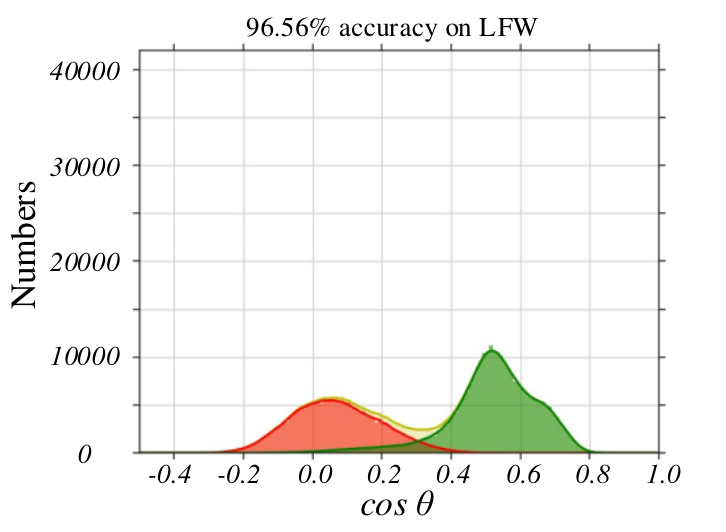}}}
\subfigure[IMDB-Face dataset]
{{\includegraphics[width=0.45\columnwidth]{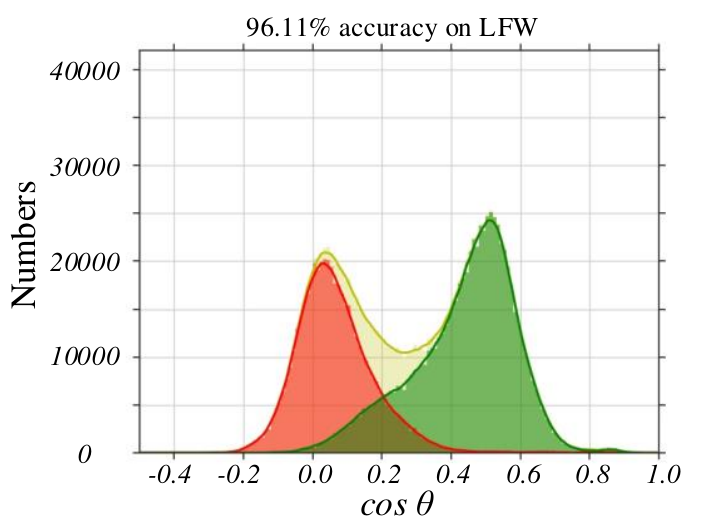}}}
\caption{The final $cos\theta$ distributions of four models (corresponding to four models Figure~\ref{figDist2}) using our paradigm.}
\label{figDist5}
\end{figure}

Corresponding to 4 models in Figure~\ref{figDist2}, we re-train them using the proposed paradigm, and the final distributions are shown in Figure~\ref{figDist5}. Our method also gets better results.

We perform experiments with different noise rates, supervised AM-Losses and computing weighted loss methods(see Section~\ref{secWL}) with the experiment in the Section~\ref{subsecEffectofNoise}. The models are evaluated on Labelled Faces in the Wild (LFW)~\cite{lfw}, Celebrities in Frontal Profile (CFP)~\cite{cfp}, and Age Database (AgeDB)~\cite{agedb}. As shown in Table~\ref{tabExp1}, competitive performance can be achieved using our paradigm, without any prior knowledge about noise in training data. We can surprisedly see that some results of $CNN_{m2}$ are even better than the results of $CNN_{clean}$. This improvement is mainly caused by semi-hard training in the final stage. It can be seen that the 2nd method in Section~\ref{secWL} demonstrates a better performance. 

For comparison, we also implemented a recently proposed noise-robust method for image classification: Co-teaching~\cite{coteaching} ($CNN_{ct}$), which selects small-loss samples from each mini-batch. \textbf{Note} the noise rate should be pre-given in Co-teaching. The results prove that general noise-robust approaches cannot achieve satisfied performance in FR.

\begin{table*}[htb]\scriptsize
\begin{center}
\resizebox{\textwidth}{18mm}{
\begin{tabular}{|c|c|c|c|c|c|c|c|c|c|c|c|c|c|c|}
\hline
\multirow{2}{*}{Loss} & Noise & \multicolumn{3}{c|}{$CNN_{clean}$} & \multicolumn{3}{c|}{$CNN_{normal}$}& $CNN_{ct}$&$CNN_{m1}$& \multicolumn{3}{c|}{$CNN_{m2}$}& Estimated \\ \cline{3-13}
                              &           Rate          & LFW    & AgeDB   &  CFP &  LFW    & AgeDB   &  CFP & LFW & LFW &  LFW     & AgeDB   &  CFP  & Noise Rate\\
\hline
   &    0\%    &94.65&79.95&82.04&94.65&79.95&82.04&-&95.00&\textbf{96.28}&\textbf{84.05}&\textbf{87.88}&2\% \\ \cline{2-14}
 L2- &    20\%   &94.18&79.33&81.00&89.05&66.83&71.55&92.12&92.95&\textbf{95.26}&\textbf{81.91}&\textbf{84.77}&18\% \\ \cline{2-14}
 Softmax  &    40\%   &92.71&76.51&77.10&85.63&58.95&68.78&87.10&89.91&\textbf{93.90}&\textbf{78.38}&\textbf{81.37}&42\% \\ \cline{2-14}
   &    60\%   &\textbf{91.15}&\textbf{70.28}&\textbf{74.74}&76.61&51.38&63.12&83.66&86.11&87.61&64.43&70.54&56\%  \\ \cline{2-14}
\hline
\multirow{4}{*}{ArcFace}  &    0\%    &97.95&88.48&\textbf{91.07}&97.95&88.48&91.07&-&97.11&\textbf{98.11}&\textbf{88.61}&90.81&2\% \\ \cline{2-14}
                          &    20\%   &\textbf{97.80}&\textbf{88.75}&89.54&96.48&82.83&82.52&96.53&96.83&97.76&88.46&\textbf{90.22}&18\% \\ \cline{2-14}
                          &    40\%   &96.53&84.93&84.81&92.33&72.68&74.11&94.25&95.88&\textbf{97.23}&\textbf{86.03}&\textbf{88.41}&36\% \\ \cline{2-14}
                          &    60\%   &94.56&80.75&80.52&84.05&58.73&67.70&90.36&93.66&\textbf{95.15}&\textbf{81.45}&\textbf{83.25}&54\% \\ \cline{2-14}
\hline
\end{tabular}
}
\end{center}
\caption{Comparison of accuracies(\%) on LFW, AgeDB(30), and CFP(FP). ResNet-20 models are used. $CNN_{clean}$ is trained only with clean data (WebFace-Clean-Sub) as \emph{Upper Bound}. $CNN_{normal}$ is trained with the noisy dataset WebFace-All using the traditional method. $CNN_{ct}$ is trained with WebFace-All using our implemented Co-teaching(with pre-given noise rates). $CNN_{m1}$ and $CNN_{m2}$ are all trained with WebFace-All but using the proposed approach, and they respectively use the 1st and 2nd method to compute loss(see Section~\ref{secWL}). $CNN_{m1}$ and $CNN_{ct}$ are only evaluated on LFW.}
\label{tabExp1}
\end{table*}

\subsection{Estimating Noise Rate}
\label{seqEst}
There is an interesting observation from Figure~\ref{figDist4} and Figure~\ref{figDist5}: at the end of training process, the region on the left of $\mu_l$ approximately contains half of noisy samples, so we can estimate the noise rate in the training dataset. If the left peak ($\mu_l$) is not detected, the region on the right of $\mu_r$, which contains about half of clean samples, also can be used. The estimated rates in Table~\ref{tabExp1} further prove the effectiveness of our method.

\subsection{Learning from Original MS-Celeb-1M}
\begin{figure}[tbh]
\centering
{{\includegraphics[width=0.9\columnwidth]{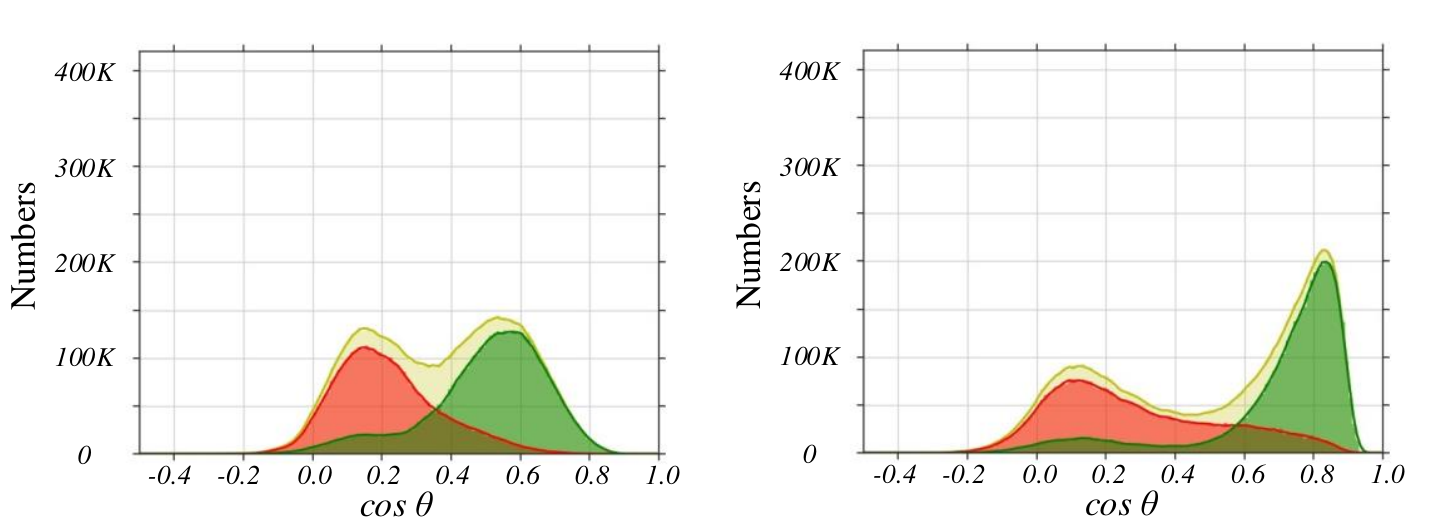}}}
\caption{The final $cos\theta$ distributions of $CNN_{normal}$ (left) and $CNN_{ours}$ (right).}
\label{figDist6}
\end{figure}

The original MS-Celeb-1M~\cite{msceleb} contains 99,892 celebrities, and 8,456,240 images. For comparison, two ResNet-64~\cite{sphereface}, $CNN_{ours}$ and $CNN_{normal}$, supervised with ArcFace~\cite{arcface} are employed to learn face features from MS-Celeb-1M, one using the proposed paradigm and the other not. To accelerate convergence speed, these ResNet-64 are firstly trained with Casia-WebFace~\cite{casia}, then finetuned with MS-Celeb-1M. Other training parameters are similar with the previous experiments. A refined MS-Celeb-1M, containing 79,077 celebrities and 4,086,798 images, is provided in LightCNN~\cite{lightcnn}, so the noise rate is about 51.6\%. We also train a CNN ($CNN_{clean}$) with the refined MS-Celeb-1M for comparison. $Hist_{all}$, together with $Hist_{clean}$ and $Hist_{noisy}$ approximated according to the refined dataset, are presented in Figure~\ref{figDist6}. According to $Hist_{noisy}$, we estimate that the noise rate of the original MS-Celeb-1M is about 43\%.

The trained CNNs are then evaluated on LFW, AgeDB-30, CFP-FP, YTF~\cite{ytf} and MegaFace Challenge 1~\cite{megaface}, as shown in Table~\ref{tabComp}. CosFace~\cite{cosface} and ArcFace~\cite{arcface} are added for comparison since they use the same network or AM-Loss with ours. The competitive performance of $CNN_{ours}$ demonstrates the effectiveness of our training paradigm.

\begin{table}[htb]\footnotesize
\begin{center}
\begin{tabular}{|c|c|c|c|c|c|}
\hline
Method & LFW & AgeDB & CFP  & YTF & MF1\\
\hline
CosFace~\cite{cosface} &99.73&-&-&97.6&77.11\\
ArcFace~\cite{arcface} &99.83&98.08&96.82&-&83.27\\
\hline
$CNN_{normal}$ &99.21&90.85&93.38&95.64&70.16\\
$CNN_{clean}$ &99.67&96.50&95.74&97.12&78.21\\
$CNN_{ours}$ &99.72&96.70&96.40&97.36&78.69\\
\hline
\end{tabular}
\end{center}
\caption{Comparison of accuracies(\%) on several public benchmarks. Accuracies of CosFace and ArcFace are cited from their original papers. CosFace is trained with a clean dataset containing 90K identities and 5M images. ArcFace is trained with a manually refined dataset containing 93K identities and 6.9M images. Their datasets all are composed of several public datasets including refined VGG2~\cite{vgg2}, MS-Celeb-1M, etc. $CNN_{ours}$ and $CNN_{normal}$ are trained only with the original noisy MS-Celeb-1M(\textbf{noise rate $\approx50\%$}). $CNN_{clean}$ is trained with the refined MS-Celeb-1M~\cite{lightcnn}. An improved ResNet-100 is used in ArcFace, and other 4 methods all use a ResNet-64 CNN model.}
\label{tabComp}
\end{table}

\section{Conclusion and Future Work}
In this paper, we propose a FR training paradigm, which employs the idea of weighting training samples, to train AM-Loss supervised CNNs with large-scale noisy data. At different stages of training process, our paradigm adjusts the weight of a sample based on the $cos\theta$ distribution to improve the robustness of the trained CNN models. Experiments demonstrate the effectiveness of our approach. Without any prior knowledge of noise, the CNN model can be directly trained with an extremely noisy dataset ($>50\%$ noisy samples), and achieves comparable performance with the model trained with an equal-size clean dataset. Moreover, the noise rate of a FR dataset can also be approximated with our approach.

The proposed paradigm also has its limitations. Firstly, it shares the same limitations with most of noise-robust training methods: the hard clean samples also have small weight, which might affect the performance. However, reducing effects of noisy samples should have higher priority while learning with a heavy noisy dataset. Secondly, Guassian-like distributions cannot be guaranteed throughout the whole training process. Fortunately, our method to find left/right peaks and end points does not heavily depend on this assumption. Lastly, we need to study the reason that the 2nd method is superior to the 1st method in Section~\ref{secWL}.

To conclude, this work will greatly reduce the requirement for clean datasets when training FR CNN models, and makes constructing huge-scale noisy FR datasets a valuable job. Moreover, our approach also can be employed to help refine a large-scale noisy FR dataset.

{\small
\bibliographystyle{ieee}
\bibliography{egbib}
}

\end{document}